\documentclass[11pt, a4paper]{thuc3i}
\usepackage[sort&compress]{natbib}
\setheadertext{JustRL: Scaling a 1.5B LLM with a Simple RL Recipe}

\usepackage[utf8]{inputenc} 
\usepackage[T1]{fontenc}    
\usepackage{hyperref}       
\usepackage{url}            
\usepackage{tabularx}
\usepackage{makecell}
\usepackage{array,ragged2e,booktabs}
\newcolumntype{P}[1]{>{\RaggedRight\arraybackslash}p{#1}}
\usepackage{longtable}
\usepackage{amsfonts}       
\usepackage{amsthm}         

\usepackage{nicefrac}       
\usepackage{microtype}      
\usepackage{xcolor}         

\usepackage{algorithm}
\usepackage{algorithmicx}
\usepackage{algpseudocode}

\definecolor{darkblue}{rgb}{0, 0, 0.5}
\hypersetup{colorlinks=true, citecolor=darkblue, linkcolor=darkblue, urlcolor=darkblue}

\usepackage{xurl}

\usepackage{soul}
\usepackage{amsmath}
\usepackage{subcaption}
\usepackage{graphicx}
\usepackage{changes}
\usepackage{mathtools}
\usepackage{changes}
\usepackage{pifont}
\usepackage{tocloft}

\usepackage{lipsum}
\usepackage{colortbl}
\usepackage{wrapfig}
\usepackage{xspace}
\usepackage{multirow}
\usepackage{enumitem}

\usepackage{pifont}
\usepackage{listings}
\usepackage{fontawesome5} 

\usepackage{wrapfig}
\usepackage{caption}

\usepackage{makecell}

\usepackage{tabularx}
\usepackage{afterpage}

\usepackage[edges]{forest}
\usepackage[normalem]{ulem}
\usepackage{caption}
\usepackage{CJKutf8}
\usepackage{bbding}
\usepackage[most,skins,theorems]{tcolorbox}
\usepackage{epigraph}

\usepackage[tikz]{bclogo}
\usepackage[framemethod=tikz]{mdframed}

\definecolor{darkblue}{rgb}{0, 0, 0.5}
\hypersetup{colorlinks=true, citecolor=darkblue, linkcolor=darkblue, urlcolor=darkblue}

\definecolor{lightorange}{HTML}{faa755}
\definecolor{lightblue}{RGB}{220,235,250}
\usepackage[most,skins,theorems]{tcolorbox}
\tcbset{
  takeawaysbox/.style={
    title=Takeaways,
    colback=lightblue!80,
    colframe=black,
    fonttitle=\bfseries\small,
    coltitle=white,
    colbacktitle=black,
    enhanced,
    attach boxed title to top left={xshift=2.5mm,yshift=-2.5mm},
    boxed title style={rounded corners, size=small, colframe=black, colback=black},
    width=\linewidth,
    arc=3.5mm
  }
}
\tcbset{
  taskbox/.style={
    title=Task Definition,
    colback=lightblue!80,
    colframe=black,
    fonttitle=\bfseries\small,
    coltitle=white,
    colbacktitle=black,
    enhanced,
    attach boxed title to top left={xshift=2.5mm,yshift=-2.5mm},
    boxed title style={rounded corners, size=small, colframe=black, colback=black},
    width=\linewidth,
    arc=2mm
  }
}
\usepackage{xcolor}

\usepackage{xspace}

\usepackage{cleveref}

\usepackage{enumitem}

\newcommand{\greencheck}{{\color{Green}\CheckmarkBold}}
\newcommand{\redcross}{{\color{Red}\XSolidBrush}}

\usepackage[dvipsnames]{xcolor}

\def\huggingface{\raisebox{-1.5pt}{\includegraphics[height=1.05em]{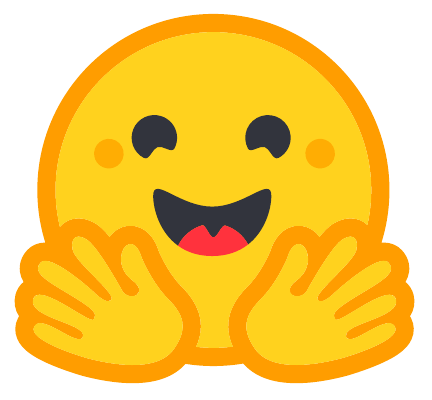}}}
\def\github{\raisebox{-1.5pt}{\includegraphics[height=1.05em]{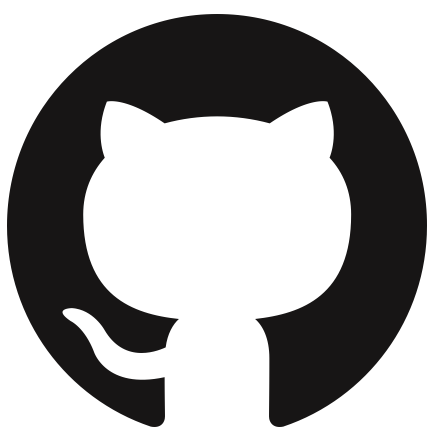}}}

\title{JustRL: Scaling a 1.5B LLM with a Simple RL Recipe}

\author{%
    Bingxiang He$^{1}$, Zekai Qu$^{1}$\hspace{0.35em}, Zeyuan Liu$^{1}$\hspace{0.4em}, Yinghao Chen$^{1}$, Yuxin Zuo$^{1}$, Cheng Qian$^{2}$, Kaiyan Zhang$^{1}$, Weize Chen$^{1}$, Chaojun Xiao$^{1}$, Ganqu Cui$^{3}$, Ning Ding$^{1\dagger}$, Zhiyuan Liu$^{1\dagger}$ \\
    $^{1}$Tsinghua University \quad
    $^{2}$University of Illinois Urbana-Champaign \quad
    $^{3}$Shanghai AI Lab \\
    \vskip1mm
    \textbf{$^\dagger$Corresponding Authors.} \quad
    \faEnvelope[regular]~\texttt{hebx24@mails.tsinghua.edu.cn}  \\
    \vskip2mm
    \huggingface \quad \url{https://huggingface.co/collections/hbx/justrl} \\
    \vskip1mm
    \github \quad \url{https://github.com/thunlp/JustRL}
}

\begin{abstract}
Recent advances in reinforcement learning for large language models have converged on increasing complexity: multi-stage training pipelines, dynamic hyperparameter schedules, and curriculum learning strategies. This raises a fundamental question: \textbf{Is this complexity necessary?} We present \textbf{JustRL}, a minimal approach using single-stage training with fixed hyperparameters that achieves state-of-the-art performance on two 1.5B reasoning models (54.9\% and 64.3\% average accuracy across nine mathematical benchmarks) while using 2$\times$ less compute than sophisticated approaches. The same hyperparameters transfer across both models without tuning, and training exhibits smooth, monotonic improvement over 4,000+ steps without the collapses or plateaus that typically motivate interventions. Critically, ablations reveal that adding ``standard tricks'' like explicit length penalties and robust verifiers may degrade performance by collapsing exploration. These results suggest that the field may be adding complexity to solve problems that disappear with a stable, scaled-up baseline. We release our models and code to establish a simple, validated baseline for the community.
\end{abstract}

\begin{document}

\maketitle

\vspace{1em}
\noindent\makebox[\textwidth]{
\parbox{0.8\textwidth}{\centering
\textit{``Perfection is achieved, not when there is nothing more to add, \\ but when there is nothing left to take away.''}\\[0.5em]
\hfill --- Antoine de Saint-Exupéry, Airman's Odyssey
}}
\vspace{1em}

\begin{figure}[h]
    \centering
    \includegraphics[width=\linewidth]{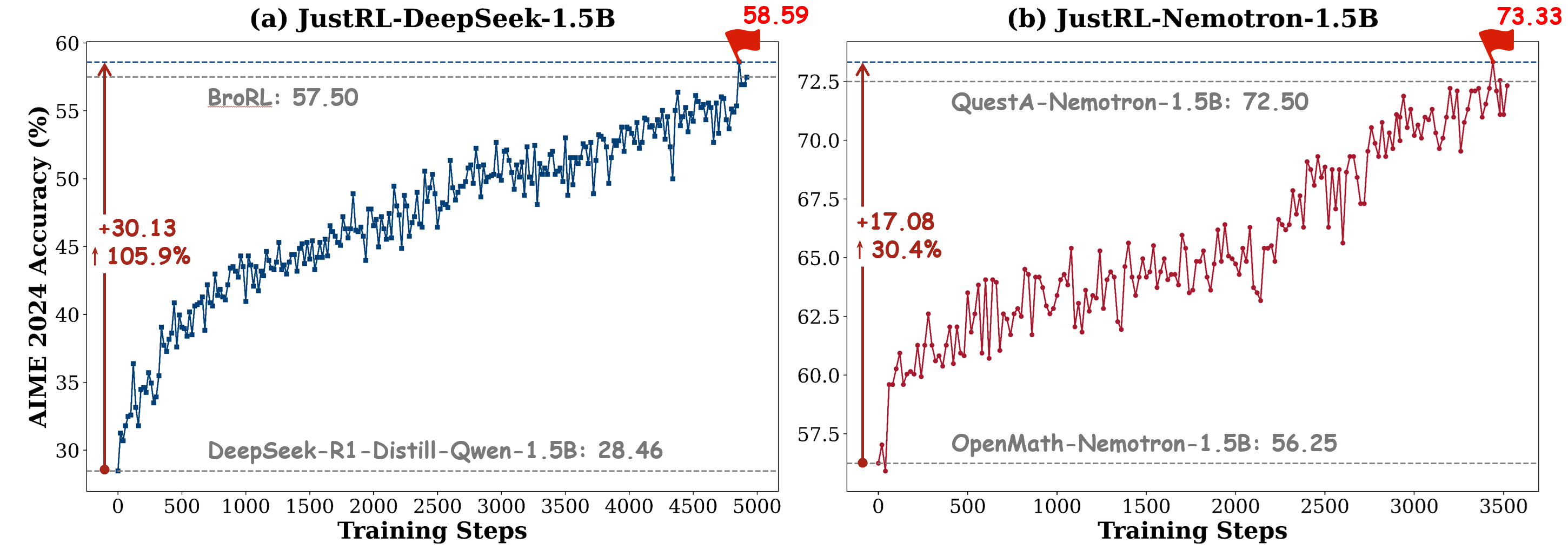}
    \caption{JustRL achieves substantial performance gains through simple, single-stage training. (a) The AIME24 (avg@32) performance curve for scaling from DeepSeek-R1-Distill-Qwen-1.5B into JustRL-DeepSeek-1.5B, from 28\% to 58\% over 4,000 steps; (b) from OpenMath-Nemotron-1.5B into our 1.5B reasoning SOTA model JustRL-Nemotron-1.5B, showing its training journey to the final 70+\% score over 3,000 steps.}
    \label{fig:aime24_scaling}
\end{figure}

\newpage

\section{Introduction}

Recent advances in Large Language Models (LLMs), such as OpenAI's o1~\citep{jaech2024openai} and DeepSeek-R1~\citep{guo2025deepseek}, have demonstrated the remarkable effectiveness of large-scale Reinforcement Learning with Verifiable Rewards (RLVR) for challenging reasoning tasks in mathematics and coding. However, for smaller lightweight models, the field has taken a different path. Leading companies have favored distillation, essentially supervised fine-tuning on outputs from larger teacher models, over direct RL training. This approach makes practical sense: distillation is efficient, stable, and delivers immediate performance gains. Qwen3's strong-to-weak distillation and DeepSeek-R1 both demonstrate the effectiveness of this strategy for small language models (SLMs).

But distillation has a fundamental limitation: it's bounded by the teacher model's capabilities. When researchers rely on distillation to improve the performance of smaller models, they encounter an upper bound, especially when the teacher model's updates are infrequent. Even with increased data and extended training, further gains in performance become difficult to achieve once the teacher model's performance plateaus. In contrast, RL can provide further improvements once the distillation process reaches saturation, making it a crucial approach in such scenarios. Meanwhile, RL for SLMs has gained a reputation for being unstable and difficult, requiring increasingly sophisticated techniques to work reliably. Over the past year, we've seen a proliferation of methods attempting to stabilize and improve RL training for small models: multi-stage training pipelines, dynamic hyperparameter schedules, adaptive temperature controls, response length penalties, and various forms of data curation and filtering~\citep{li2025questa,min2024imitate,deepscaler2025,liu2025prorl,hu2025prorlv2,hu2025brorl}.

This proliferation of techniques raises an important question: \textbf{Is this complexity necessary?} When different works combine different subsets of methods and report varying results, it becomes unclear what truly drives performance. More concerning, many recent works cite training instabilities, like reward collapse, entropy drift, and length explosion, as motivation for their techniques, yet apply these techniques on top of already-complex baselines. This makes it impossible to know whether new methods provide genuine benefits or simply compensate for issues introduced by prior complexity. The accumulated ``best practices'' may be fighting each other rather than the fundamental challenges of RL~\citep{liu2025part}.

In this paper, we explore \textbf{whether stable, competitive training can be achieved with a simpler approach.} We apply a minimal setup to two popular 1.5B reasoning models, using single-stage training with fixed hyperparameters derived from common practice. The results match or exceed more complex approaches while using 2$\times$ less compute. Importantly, we achieve this without the multi-stage pipelines or dynamic schedules, suggesting that simpler approaches may be sufficient when applied at adequate scale. Besides, the training process itself proves stable: smooth, monotonic improvement over 4,000+ steps without the collapses or oscillations often cited as motivation for complex interventions.

Our goal is not to argue against all techniques or claim we've found the optimal approach. Rather, we provide evidence that simpler baselines deserve more attention than they've received. We offer a simple practice with a minimum set of tricks that can enhance the performance of models that are approaching their distillation limits. The field may benefit from establishing what's fundamentally sufficient before layering on additional complexity.

\section{Related Work}

Since DeepSeek-R1's release in early 2025, the community has rapidly advanced RL for small language models in mathematical reasoning. The past year has seen a flourishing of approaches, each introducing techniques to stabilize training and push performance boundaries. These works fall into three main families based on their foundation models, all starting from distilled bases: (1) DeepSeek-R1-Distill-Qwen-1.5B, (2) OpenMath-Nemotron-1.5B, and (3) Qwen3-1.7B.

The evolution reveals a clear trend toward increasing sophistication. Early works like STILL~\citep{min2024imitate} explored hyperparameter tuning and reference model resets through extensive comparison experiments. Subsequent approaches introduced multi-stage training with progressive context lengthening. DeepScaleR~\citep{deepscaler2025} divided training into three stages with increasing context lengths (8K $\rightarrow$ 16K $\rightarrow$ 24K). FastCuRL~\citep{song2025fastcurl} extended this to five stages, alternating between CoT compression (long-to-short) and extension (short-to-long), with each stage using different data, batch sizes, and rollout numbers. ProRL~\citep{liu2025prorl} divided training into eight stages with scheduled length penalties, and its successor ProRL-V2~\citep{hu2025prorlv2} introduced additional techniques including scheduled cosine length penalties while maintaining fixed 8K context. BroRL~\citep{hu2025brorl} took a different approach by dramatically increasing rollouts per example to hundreds, aiming to exhaustively explore the solution space after 3K ProRL training steps.

\begin{table*}[t]
\centering
\scriptsize
\renewcommand{\arraystretch}{1.5}
\resizebox{\textwidth}{!}{
\begin{tabular}{lcccccccccc}
\toprule
\textbf{Model} & \textbf{EC} & \textbf{THP} & \textbf{TTP} & \textbf{RKL} & \textbf{LC} & \textbf{AT} & \textbf{RR} & \textbf{DS} & \textbf{ST} & \textbf{Date} \\
\midrule
\rowcolor[HTML]{EEF6C6}STILL-3-1.5B & \redcross & \greencheck & \greencheck & \greencheck & \redcross & \redcross & \redcross & \redcross & \redcross & Jan '25 \\
\rowcolor[HTML]{EEF6C6}DeepScaleR-1.5B & \greencheck & \redcross & \redcross & \redcross & \greencheck & \redcross & \redcross & \redcross & \greencheck & Feb '25 \\
\rowcolor[HTML]{EEF6C6}FastCuRL-1.5B & \redcross & \greencheck & \redcross & \redcross & \greencheck & \redcross & \redcross & \redcross & \greencheck & Mar '25 \\
\rowcolor[HTML]{EEF6C6}ProRL-V1 & \greencheck & \greencheck & \redcross & \greencheck & \greencheck & \redcross & \redcross & \greencheck & \greencheck & May '25 \\
\rowcolor[HTML]{D9F3FD}e3-1.7B & \greencheck & \greencheck & \redcross & \redcross & \greencheck & \redcross & \redcross & \greencheck & \greencheck & Jun '25 \\
\rowcolor[HTML]{D9F3FD}POLARIS-1.7B & \greencheck & \greencheck & \redcross & \redcross & \greencheck & \greencheck & \greencheck & \greencheck & \greencheck & Jul '25 \\
\rowcolor[HTML]{EEF6C6}ProRL-V2 & \greencheck & \greencheck & \redcross & \greencheck & \greencheck & \redcross & \redcross & \greencheck & \greencheck & Aug '25 \\
\rowcolor[HTML]{FBBFBC}QuestA-Nemotron & \redcross & \redcross & \redcross & \redcross & \redcross & \redcross & \redcross & \greencheck & \greencheck & Sep '25 \\
\rowcolor[HTML]{EEF6C6}BroRL & \greencheck & \greencheck & \redcross & \greencheck & \greencheck & \redcross & \redcross & \greencheck & \greencheck & Oct '25 \\
\midrule
\rowcolor[HTML]{EEF6C6}\textbf{JustRL-DeepSeek} & \greencheck & \redcross & \redcross & \redcross & \redcross & \redcross & \redcross & \redcross & \redcross & Nov '25 \\
\rowcolor[HTML]{FBBFBC}\textbf{JustRL-Nemotron} & \greencheck & \redcross & \redcross & \redcross & \redcross & \redcross & \redcross & \redcross & \redcross & Nov '25 \\
\bottomrule
\end{tabular}}
\caption{Comparison of RL techniques used in recent small language models for mathematical reasoning. Model names are colored by backbone: \colorbox[HTML]{EEF6C6}{DeepSeek-R1-Distill-Qwen-1.5B}, \colorbox[HTML]{D9F3FD}{Qwen3-1.7B}, \colorbox[HTML]{FBBFBC}{OpenMath-Nemotron-1.5B}. We use the following abbreviations for RL techniques: EC=Entropy Control, THP=Tune Hyperparameters, TTP=Tune Training Prompt, RKL=Reset KL Reference, LC=Length Control, AT=Adaptive Temperature, RR=Rollout Rescue, DS=Dynamic Sampling, ST=Split Training Stages. Our models (JustRL-DeepSeek and JustRL-Nemotron) use only entropy control, achieving competitive performance with minimal complexity.}
\label{tab:landscape}
\end{table*}

For the OpenMath-Nemotron-1.5B backbone, QuestA~\citep{li2025questa} introduced an innovative curriculum learning approach by augmenting questions with partial CoT solutions as hints, providing richer learning signals through staged difficulty progression. On the Qwen3-1.7B backbone, POLARIS~\citep{Polaris2025} employed dynamic dataset filtering to focus on challenging problems, combined with adaptive temperature adjustments and test-time context extrapolation across three training stages. Similarly, e3~\citep{setlur2025e3} used multi-stage training with varying context lengths and leveraged the model's extrapolation abilities at test time.

\Cref{tab:landscape} summarizes these approaches and the techniques they employ. The pattern is striking: nearly every work employs multiple techniques from a growing toolkit, including multi-stage training, adaptive hyperparameters, length penalties, dynamic sampling, and various stabilization mechanisms. While these methods achieve strong results, the accumulated complexity makes it difficult to isolate which elements truly matter. This raises a practical question: Is there a simpler path that still achieves competitive performance?

\section{JustRL: Simplicity at Scale}

Our approach is deliberately simple. We constrain ourselves to the fundamentals of RL, avoiding the multi-stage pipelines, dynamic schedules, and specialized techniques that have become common in recent work.

\subsection{Training Setup}

\textbf{Core algorithm.} We use default implementation of GRPO in veRL~\citep{sheng2025hybridflow} with binary outcome rewards. The reward signal comes from a lightweight rule-based verifier from DAPO~\citep{yu2025dapo}, without symbolic math libraries like SymPy that could add computational overhead.

\textbf{What we keep simple:}
\begin{itemize}[topsep=0pt, itemsep=0pt, leftmargin=18pt]
\item \textbf{Single-stage training:} No progressive context lengthening, no curriculum switching, no stage transitions. We train continuously from start to finish.
\item \textbf{Fixed hyperparameters:} No adaptive temperature scheduling, no dynamic batch size adjustments, no mid-training reference model resets.
\item \textbf{Standard data:} We train on DAPO-Math-17k~\citep{yu2025dapo} without offline difficulty filtering or online dynamic sampling strategies.
\item \textbf{Basic prompting:} A simple suffix prompt without tuning: ``Please reason step by step, and put your final answer within $\backslash$boxed\{\}.''
\item \textbf{Length control:} We simply set the maximum context length as 16K tokens, rather than using explicit length penalty terms.
\end{itemize}

\textbf{The one technique we do use.} We employ ``clip higher'', a well-established practice for stability in long-horizon RL training. We view this as part of the baseline rather than an added technique.

\begin{table}[t]
\centering
\small
\begin{tabular}{lc}
\toprule
\textbf{Hyperparameter} & \textbf{Value} \\
\midrule
Advantage Estimator & GRPO \\
Use KL Loss & No \\
Use Entropy Regularization & No \\
Train Batch Size & 256 \\
Max Prompt Length & 1k \\
Max Response Length & 15k \\
PPO Mini Batch Size & 64 \\
PPO Micro Batch Size/GPU & 1 \\
Clip Ratio Range & [0.8, 1.28] \\
Learning Rate & 1e-6 (constant) \\
Temperature & 1.0 \\
Rollout N & 8 \\
Reward Function & DAPO~\citep{yu2025dapo} \\
\bottomrule
\end{tabular}
\caption{Fixed hyperparameter configuration used for both JustRL models.}
\label{tab:hyperparams}
\end{table}

We train this recipe on two 1.5B reasoning models using veRL: DeepSeek-R1-Distill-Qwen-1.5B and OpenMath-Nemotron-1.5B, each with 32 A800-80GB GPUs for $\sim$15 days. The same hyperparameters work for both, without per-model tuning, and remain fixed throughout training. \Cref{tab:hyperparams} shows the complete hyperparameter configuration.

\subsection{Evaluation Protocol}

We evaluate nine challenging mathematical reasoning tasks based on reproducible evaluation scripts from POLARIS~\citep{Polaris2025}:

\begin{itemize}[topsep=0pt, itemsep=0pt, leftmargin=18pt]
\item \textbf{Benchmarks:} AIME 2024~\citep{li2024numinamath}, AIME 2025~\citep{balunovic2025matharena}, AMC 2023~\citep{li2024numinamath}, MATH-500~\citep{hendrycks2021measuring}, Minerva Math~\citep{lewkowycz2022solving}, OlympiadBench~\citep{he-etal-2024-olympiadbench}, HMMT Feb 2025~\citep{balunovic2025matharena}, CMIMC 2025~\citep{balunovic2025matharena} and BRUMO 2025~\citep{balunovic2025matharena}.
\item \textbf{Evaluation protocol:} We report Pass@1 accuracy, averaging over N sampled responses per problem (N=4 for MATH-500, Minerva Math, and OlympiadBench; N=32 for others). We use temperature 0.7, top-p 0.9, and allow up to 32K tokens for generation.
\end{itemize}

We augment existing systems with CompassVerifier-3B~\citep{CompassVerifier}, a lightweight model-based verifier, to address false negatives from rule-based verifiers.

\section{Experimental Results}

We apply JustRL on two popular 1.5B reasoning models to demonstrate that our minimal recipe achieves competitive performance with notably stable training dynamics.

\subsection{Scaling a Weaker Base: JustRL-DeepSeek-1.5B}

\begin{tcolorbox}[title = \textbf{Takeaway 1}, colback=Salmon!20, colframe=Salmon!90!Black]

Starting from DeepSeek-R1-Distill-Qwen-1.5B, we achieve better results through single-stage training with fixed hyperparameters, outperforming more complex approaches while using 2× less compute. The training curve shows over 4,000 steps of stable improvement without intervention, suggesting that an adequate scale with simple methods can outperform sophisticated techniques.

\end{tcolorbox}

We train DeepSeek-R1-Distill-Qwen-1.5B for 4,380 steps using our simple, single-stage recipe. We report the avg@32 results across nine mathematical benchmarks in \Cref{tab:deepseek_results}.

\textbf{Results.} Our model (JustRL-DeepSeek-1.5B) achieves 54.87\% average across benchmarks, outperforming ProRL-V2's 53.08\% despite ProRL-V2's nine-stage training pipeline with dynamic hyperparameters and more sophisticated techniques. We also lead on six of nine benchmarks, demonstrating broad improvements rather than overfitting to a single task.

\begin{table*}[t]
\centering
\small
\setlength{\tabcolsep}{2pt}
\begin{tabular}{lcccccccccc}
\toprule
\textbf{Model} & \textbf{AIME24} & \textbf{AIME25} & \textbf{AMC23} & \textbf{MATH} & \textbf{Minerva} & \textbf{Olympiad} & \textbf{HMMT} & \textbf{BRUMO} & \textbf{CMIMC} & \textbf{Avg} \\
\midrule
Backbone & 29.90 & 22.40 & 63.82 & 84.90 & 34.65 & 45.95 & 13.44 & 30.94 & 12.89 & 37.65 \\
DeepScaleR-1.5B & 40.21 & 28.65 & 73.83 & 89.30 & 39.34 & 52.79 & 18.96 & 40.00 & 21.00 & 44.88 \\
ProRL-V2 & 51.87 & 35.73 & \underline{88.75} & \underline{92.00} & 49.03 & \underline{67.84} & \underline{19.38} & \underline{47.29} & \textbf{25.86} & \underline{53.08} \\
BroRL$^*$ & \textbf{57.50} & \underline{36.88} & -- & \textbf{92.14} & \underline{49.08} & 61.54 & -- & -- & -- & -- \\
\midrule
\textbf{JustRL-DeepSeek} & \underline{52.60} & \textbf{38.75} & \textbf{91.02} & 91.65 & \textbf{51.47} & \textbf{67.99} & \textbf{21.98} & \textbf{52.71} & \underline{25.63} & \textbf{54.87} \\
\bottomrule
\end{tabular}
\caption{Results on DeepSeek-R1-Distill-Qwen-1.5B backbone. All scores except MATH-500, Minerva, and OlympiadBench use @32 sampling; those three use @4. $^*$BroRL results are officially reported but models not released; some benchmarks unavailable.}
\label{tab:deepseek_results}
\end{table*}

\textbf{Computational efficiency.} However, the real question is whether our simplicity comes at a computational cost. It doesn't. \Cref{tab:compute} compares the computational cost across methods. We match half of ProRL-V2's compute budget while using a single-stage recipe with fixed hyperparameters. BroRL requires 4.9$\times$ more compute by increasing rollouts to 512 per example, essentially exhaustively exploring the solution space. Our approach achieves competitive performance without this computational overhead.

\begin{table*}[t]
\centering
\small
\setlength{\tabcolsep}{3.5pt}
\renewcommand{\arraystretch}{1.2}
\begin{tabular}{@{}lcccccc@{}}
\toprule
\textbf{Model} & \makecell{\textbf{Dynamic}\\\textbf{Sampling$^*$}} & \makecell{\textbf{Training}\\\textbf{Steps}} & \makecell{\textbf{Train}\\\textbf{Batch Size}} & \makecell{\textbf{Rollout N}} & \makecell{\textbf{Max Context}\\\textbf{Length}} & \makecell{\textbf{Token Budget}\\\textbf{(approx.)}} \\
\midrule
DeepScaleR-1.5B & \redcross & 1,750 & 128 & 8 & 8k$\to$16k$\to$24k & 2.2$\times$10$^6$k \\
ProRL-V1 & \greencheck & 2,450 & 256 & 16$\to$32$\to$16 & 8k$\to$16k & 2.1$\times$10$^8$k \\
ProRL-V2 & \greencheck & +1,000 & 256 & 16$\to$32$\to$16 & 8k$\to$16k$\to$8k & 2.8$\times$10$^8$k \\
BroRL & \greencheck & +191 & 128 & 512 & 16k & 6.8$\times$10$^8$k \\
\midrule
\textbf{JustRL-DeepSeek} & \redcross & 4,380 & 256 & 8 & 16k & 1.4$\times$10$^8$k \\
\bottomrule
\end{tabular}
\caption{Computational cost comparison for DeepSeek-R1-Distill-Qwen-1.5B based models. $^*$Dynamic sampling with estimated 50\% filter ratio following POLARIS~\citep{Polaris2025}. ProRL-V2 continues from ProRL-V1 (+1,000 steps), and BroRL continues from ProRL-V2 (+191 steps).}
\label{tab:compute}
\end{table*}

\textbf{Note on dynamic sampling.} Models marked with \greencheck use dynamic sampling to filter examples. Following POLARIS~\citep{Polaris2025}, we estimate a 50\% filter ratio for DeepSeek-R1-Distill-Qwen-1.5B using dynamic sampling, as rollouts often contain many trivial/hard cases (e.g., 8/8 or 0/8 correct rollouts). Even assuming no filtering (i.e., 0\% ratio), our compute use remains comparable or even lower, making our estimates conservative.

\textbf{Training stability.} \Cref{fig:aime24_scaling}(a) shows our training curve for JustRL-DeepSeek-1.5B, showing smooth and monotonic improvement without the oscillations or plateaus that typically require intervention. The stability itself suggests we're not fighting against our training setup.

\subsection{Scaling a Stronger Base: JustRL-Nemotron-1.5B}

\begin{tcolorbox}[title = \textbf{Takeaway 2}, colback=Salmon!20, colframe=Salmon!90!Black]

The same recipe scales OpenMath-Nemotron-1.5B to the current best math reasoning performance without any hyperparameter adjustment, matching state-of-the-art results that use curriculum learning and question augmentation. Competitive performance across two different starting points suggests the approach is robust rather than carefully tuned to specific conditions.

\end{tcolorbox}

We train OpenMath-Nemotron-1.5B for 3,440 steps using the identical recipe, without hyperparameter changes. We report the evaluation results across nine challenging mathematical benchmarks in \Cref{tab:nemotron_results}.

\begin{table*}[t]
\centering
\small
\setlength{\tabcolsep}{2pt}
\begin{tabular}{lcccccccccc}
\toprule
\textbf{Model} & \textbf{AIME24} & \textbf{AIME25} & \textbf{AMC23} & \textbf{MATH} & \textbf{Minerva} & \textbf{Olympiad} & \textbf{HMMT} & \textbf{BRUMO} & \textbf{CMIMC} & \textbf{Avg} \\
\midrule
Backbone & 58.75 & 48.44 & 90.55 & 92.40 & 26.93 & 71.70 & 30.10 & 61.67 & 30.08 & 56.74 \\
QuestA & \textbf{71.56} & \underline{62.08} & \underline{93.44} & \underline{92.95} & \textbf{32.08} & \underline{72.28} & \textbf{40.94} & \textbf{67.50} & \underline{41.48} & \underline{63.81} \\
\midrule
\textbf{JustRL-Nemotron} & \underline{69.69} & \textbf{62.92} & \textbf{96.02} & \textbf{94.15} & \underline{30.24} & \textbf{76.59} & \underline{40.63} & \underline{66.88} & \textbf{41.72} & \textbf{64.32} \\
\bottomrule
\end{tabular}
\caption{Results on OpenMath-Nemotron-1.5B backbone. All scores except MATH-500, Minerva, and OlympiadBench use @32 sampling; those three use @4.}
\label{tab:nemotron_results}
\end{table*}

\textbf{Results.} We achieve 64.32\% average, slightly outperforming QuestA's 63.81\% and leading on five of nine benchmarks. The gap is narrow, which makes sense. Both approaches are pushing the boundaries of what's achievable at 1.5B scale. The key difference is in how we get there.

QuestA introduces an innovative curriculum learning approach that augments questions with partial CoT solutions as hints, splitting training stages with different difficulty. This requires not just ground-truth answers but full reasoning trajectories generated by larger models for curriculum construction with additional data requirements and engineering complexity. Our approach uses only the standard question-answer pairs without augmentation or curriculum design.

\begin{table}[t]
\centering
\small
\setlength{\tabcolsep}{3.5pt}
\renewcommand{\arraystretch}{1.2}
\begin{tabular}{@{}lcccccc@{}}
\toprule
\textbf{Model} & \makecell{\textbf{Dynamic}\\\textbf{Sampling$^*$}} & \makecell{\textbf{Training}\\\textbf{Steps}} & \makecell{\textbf{Train}\\\textbf{Batch Size}} & \makecell{\textbf{Rollout N}} & \makecell{\textbf{Max Context}\\\textbf{Length}} & \makecell{\textbf{Token Budget}\\\textbf{(approx.)}} \\
\midrule
QuestA & \greencheck & 2,000 & 128 & 16 & 32k & 2.6$\times$10$^8$k \\
\midrule
\textbf{JustRL-Nemotron} & \redcross & 3,440 & 256 & 8 & 16k & 1.1$\times$10$^8$k \\
\bottomrule
\end{tabular}
\caption{Computational cost comparison for OpenMath-Nemotron-1.5B based models. $^*$Dynamic sampling with estimated 50\% filter ratio. Despite more training steps, JustRL-Nemotron uses 2.4$\times$ less compute.}
\label{tab:nemotron_compute}
\end{table}

\textbf{Computational efficiency.} We use 2$\times$ less compute while achieving slightly better average performance without designing a complex curriculum as used in QuestA.

\textbf{Training stability.} \Cref{fig:aime24_scaling}(b) shows another smooth training curve. The fact that the same recipe works for both models without hyperparameter tuning suggests genuine robustness rather than lucky optimization for a single model. 

These results don't diminish QuestA's contribution, where question augmentation is a clever technique that clearly helps. Rather, they demonstrate that competitive performance is achievable through simpler means.

\subsection{Training Dynamics Analysis}

\begin{figure*}[t]
\centering
\includegraphics[width=\linewidth]{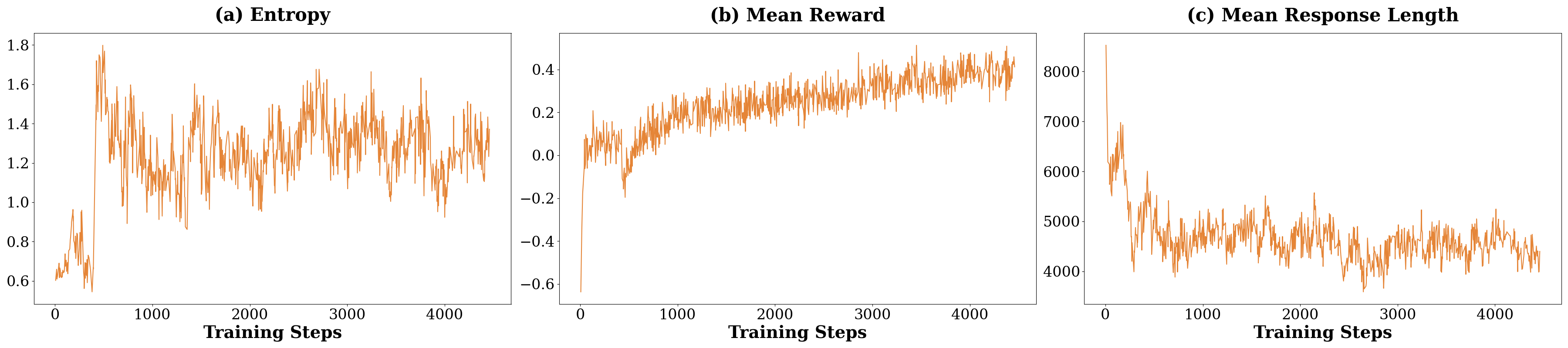}
\caption{Training Dynamics of JustRL-DeepSeek-1.5B. (a) Policy entropy remains stable throughout training, oscillating naturally around 1.2-1.4 without drift or collapse. (b) Mean reward shows smooth, monotonic improvement from negative to $\sim$0.4, indicating consistent learning without plateau-breaking interventions. (c) Response length naturally converges from initial verbosity ($\sim$7,000 tokens) to a stable range (4,000-5,000 tokens) with 16k max context length, without explicit length penalties.}
\label{fig:dynamics}
\end{figure*}

The ultimate test of a training recipe isn't just the final numbers; it's whether you can get there reliably. Complex techniques often emerge as responses to training instability: oscillating rewards, collapsing policies, or runaway response lengths. If a simpler approach can avoid these failure modes entirely, it suggests we may have been treating symptoms rather than causes. We examine the training dynamics of JustRL-DeepSeek-1.5B in detail, tracking three key dynamics over 4,000 training steps: mean training reward, policy entropy, and mean response length in \Cref{fig:dynamics}. These dynamics reveal whether the model is learning stably or requires constant intervention.

\begin{itemize}[topsep=0pt, itemsep=0pt, leftmargin=18pt]
\item \textbf{Entropy:} \Cref{fig:dynamics}(a) shows policy entropy oscillating between 1.0 and 1.6 at later training steps, with no systematic drift upward (exploration collapse) or downward (premature convergence), indicating that the simple “clip higher” technique is well-performed for large-scale RL.
\item \textbf{Mean Reward:} \Cref{fig:dynamics}(b) shows the mean reward climbing from around -0.6 to +0.4 over training. The curve is noisy but the trend is unmistakably upward. More importantly, there are no extended plateaus or sudden drops that would typically trigger intervention in multi-stage approaches. The signal is consistent enough that the model can learn continuously.
\item \textbf{Mean Response Length:} The model starts verbose, generating responses averaging $\sim$8,000 tokens. Without any explicit length penalty, it naturally compresses to 4,000-5,000 tokens by step 1,000 and maintains this range. This organic compression may be more robust than explicit penalties, which can create adversarial pressure that models learn to game~\citep{liu2025dler}.
\end{itemize}

\textbf{The contrast with typical RL:} While we don't have the computational resources to run extensive controlled comparisons, the literature provides context. Many recent works explicitly cite training instabilities as motivation for their techniques: ProRL-v2~\citep{hu2025prorlv2} introduces scheduled length penalties after observing length drift; BroRL~\citep{hu2025brorl} increases rollouts to hundreds after hitting plateaus; multiple works~\citep{liu2025prorl,min2024imitate} apply KL regularization and reset reference models when KL divergence grows too large, which limits the training upper bound. Our training exhibits none of these pathologies that motivate intervention.

\textbf{What we can't claim:} These smooth curves don't prove that simpler approaches are always more stable, or that techniques never help. We can't isolate which specific complex techniques cause instability versus which ones solve it. But the contrast is striking: a minimal recipe produces training dynamics that simply don't require the interventions that have become standard practice.

\subsection{Ablation Studies}

We conduct two ablation studies starting from our base recipe on JustRL-DeepSeek-1.5B, both trained for 3,000+ steps:

\begin{itemize}[topsep=0pt, itemsep=0pt, leftmargin=18pt]
\item \textbf{w/ Overlong Penalty:} Add an explicit length penalty term for the last 4k tokens (as used in DAPO~\citep{yu2025dapo})
\item \textbf{w/ Overlong Penalty + Robust Verifier:} Further add a more sophisticated verifier from DeepScaleR~\citep{deepscaler2025} to reduce false negatives
\end{itemize}

\begin{figure}[t]
\centering
\includegraphics[width=0.8\linewidth]{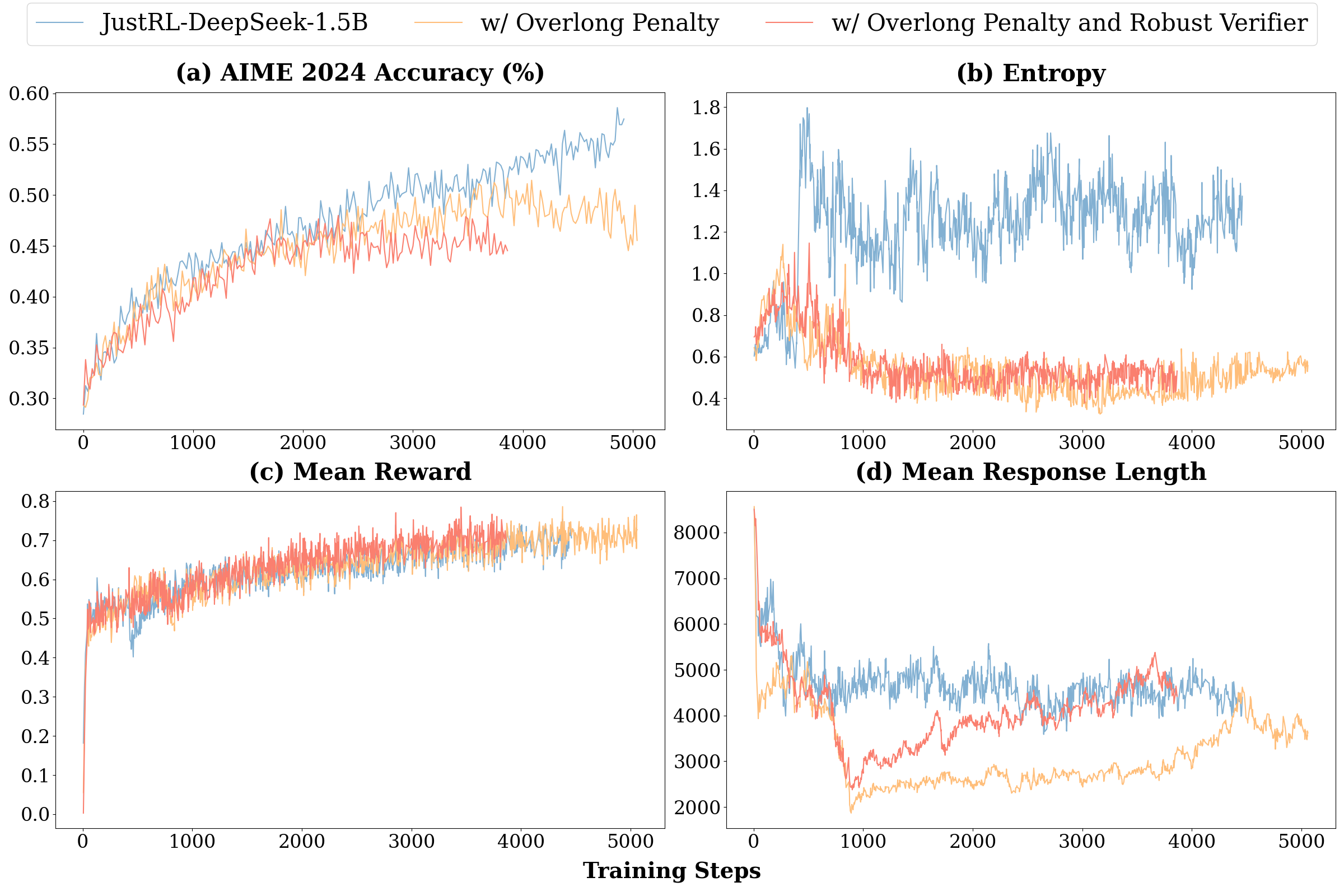}
\caption{Ablation Study Results. (a) AIME 2024 performance diverges after $\sim$2,000 steps. Our base recipe reaches 55\%, while adding overlong penalty plateaus at 50\%, and adding both modifications plateaus at 45\%. (b) Entropy: Both modifications show collapsed exploration (entropy $\sim$0.5-0.6) compared to healthy oscillation in the base recipe ($\sim$1.2-1.4).}
\label{fig:ablation}
\end{figure}

\textbf{Results.} \Cref{fig:ablation} shows that both modifications degrade performance: adding overlong penalty plateaus at 50\% AIME 2024 (vs 55\% baseline), and adding both modifications plateaus at 45\%.

\textbf{On the overlong penalty.} We hypothesized that explicitly penalizing verbose responses might improve training efficiency by pushing the model toward conciseness faster. Instead, performance degraded significantly as a trade-off. The entropy plot in \Cref{fig:ablation}(b) reveals why: the explicit penalty collapses exploration, driving entropy down to 0.5-0.6 compared to the 1.2-1.4 range in our base approach. The explicit penalty appears to create pressure that conflicts with the learning objective, forcing premature convergence to shorter responses before the model has explored what actually works.

\textbf{On the robust verifier.} We further hypothesized that reducing false negatives (correct solutions marked wrong) would provide a cleaner learning signal. However, even after normalizing reward scales, its use leads to worse final performance, plateauing at 45\% AIME 2024. We offer two possible explanations: first, the stricter base verifier creates a richer spectrum of learning signals by reducing ``perfect'' scores, whereas the robust verifier's permissiveness offers less nuanced guidance. Second, the stricter verifier's reliance on precise formatting may pressure the model to develop more robust internal computations, an incentive lost when the verifier corrects errors externally. Thus, a forgiving verifier might fail to encourage the precision required for optimal generalization.

These results reveal two important lessons. First, not all ``standard tricks'' transfer across contexts. The overlong penalty works in DAPO's setting~\citep{yu2025dapo} but degrades performance in ours, demonstrating that techniques interact with other design choices in complex and sometimes unpredictable ways. Second, simpler approaches are not always easier to improve. We tested two seemingly reasonable modifications and both made things worse, suggesting our base recipe achieves a delicate balance that is easily disrupted by additional interventions.

We want to be clear about the limits of these ablations. We tested two specific modifications, but many other techniques remain unexplored: curriculum learning, adaptive temperature scheduling, reference model resets, different verifier designs, and various forms of data augmentation. Some of these might improve upon our baseline. Our point is not that additional techniques \emph{never} help, rather, it is that they should be validated empirically rather than assumed to be beneficial.

\section{Discussion}

\textbf{What this suggests:} The smooth training curves with healthy entropy, monotonic rewards and natural length convergence stand in contrast to instabilities often cited as motivation for complex techniques. Our negative ablations show that adding ``improvements'' actively degrades performance. This suggests complexity may sometimes address symptoms created by other design choices rather than fundamental RL challenges.

\textbf{What we don't know:} We demonstrate that simple RL works well, but can't isolate why. Is it the hyperparameters? The training dataset? The verifier design? The interaction between all three? Our results are also limited to two backbones in mathematical reasoning at 1.5B scale. Generalization to other domains, model sizes, and tasks remains an open question.

\textbf{When might complexity help:} We don't advocate simplicity as dogma. Additional techniques may be valuable under extreme compute constraints, when encountering specific failure modes we didn't face, when pushing beyond current performance ceilings, or in domains with noisier reward signals. Our argument is methodological: \textbf{establish simple baselines first, then add complexity only when you identify specific problems it solves.}

\section{Conclusion}

The debate over RL for small models has been clouded by assumptions that complexity is necessary for stability and performance. We set out to answer a straightforward question: What happens if we apply RL to small language models without specialized techniques that have become standard practice? By stepping back to a simpler approach, our findings provide a clear answer: adequate scale with stable fundamentals can match sophisticated techniques. Starting from two foundation models, we achieved comparable or better performance using single-stage training with fixed hyperparameters, matching or exceeding approaches that employ multi-stage training and curriculum learning while using 2$\times$ less compute. More striking than the final numbers is the path: smooth, stable improvement over thousands of steps without the interventions typically required to prevent training collapse. We advocate a methodological shift: \textbf{start simple, scale up, and only add complexity when a simple, robust baseline demonstrably fails.} If simplicity is sufficient more often than current practice assumes, that seems worth paying attention to.

\section*{Limitations}

Our work has several limitations. First, our results are limited to mathematical reasoning tasks at the 1.5B parameter scale, and generalization to other domains (e.g., coding, general question answering) and model sizes remains unexplored. Second, while we demonstrate that simplicity works, we cannot definitively isolate which specific components (hyperparameters, verifier design, training data) are most critical to our success. Third, our compute budget, while lower than some complex methods, may still be prohibitive for resource-constrained researchers. Finally, we have not explored whether our approach maintains advantages when pushed to even longer training horizons or whether additional techniques might become necessary at scale.

\newpage

\bibliography{JustRL}

@article{jaech2024openai,
  title={Openai o1 system card},
  author={Jaech, Aaron and Kalai, Adam and Lerer, Adam and Richardson, Adam and El-Kishky, Ahmed and Low, Aiden and Helyar, Alec and Madry, Aleksander and Beutel, Alex and Carney, Alex and others},
  journal={arXiv preprint arXiv:2412.16720},
  year={2024}
}

@article{guo2025deepseek,
  title={Deepseek-r1 incentivizes reasoning in llms through reinforcement learning},
  author={Guo, Daya and Yang, Dejian and Zhang, Haowei and Song, Junxiao and Wang, Peiyi and Zhu, Qihao and Xu, Runxin and Zhang, Ruoyu and Ma, Shirong and Bi, Xiao and others},
  journal={Nature},
  volume={645},
  number={8081},
  pages={633--638},
  year={2025},
  publisher={Nature Publishing Group UK London}
}

@article{li2025questa,
  title={Questa: Expanding reasoning capacity in llms via question augmentation},
  author={Li, Jiazheng and Lin, Hongzhou and Lu, Hong and Wen, Kaiyue and Yang, Zaiwen and Gao, Jiaxuan and Wu, Yi and Zhang, Jingzhao},
  journal={arXiv preprint arXiv:2507.13266},
  year={2025}
}

@article{min2024imitate,
  title={Imitate, explore, and self-improve: A reproduction report on slow-thinking reasoning systems},
  author={Min, Yingqian and Chen, Zhipeng and Jiang, Jinhao and Chen, Jie and Deng, Jia and Hu, Yiwen and Tang, Yiru and Wang, Jiapeng and Cheng, Xiaoxue and Song, Huatong and others},
  journal={arXiv preprint arXiv:2412.09413},
  year={2024}
}

@misc{deepscaler2025,
  title={DeepScaleR: Surpassing O1-Preview with a 1.5B Model by Scaling RL},
  author={Michael Luo and Sijun Tan and Justin Wong and Xiaoxiang Shi and William Y. Tang and Manan Roongta and Colin Cai and Jeffrey Luo and Li Erran Li and Raluca Ada Popa and Ion Stoica},
  howpublished={\url{https://pretty-radio-b75.notion.site/DeepScaleR-Surpassing-O1-Preview-with-a-1-5B-Model-by-Scaling-RL-19681902c1468005bed8ca303013a4e2}},
  note={Notion Blog},
  year={2025}
}

@article{liu2025prorl,
  title={Prorl: Prolonged reinforcement learning expands reasoning boundaries in large language models},
  author={Liu, Mingjie and Diao, Shizhe and Lu, Ximing and Hu, Jian and Dong, Xin and Choi, Yejin and Kautz, Jan and Dong, Yi},
  journal={arXiv preprint arXiv:2505.24864},
  year={2025}
}

@misc{hu2025prorlv2,
  title        = {ProRL V2: Prolonged Training Validates RL Scaling Laws},
  author       = {Jian Hu and Mingjie Liu and Shizhe Diao and Ximing Lu and Xin Dong and Pavlo Molchanov and Yejin Choi and Jan Kautz and Yi Dong},
  year         = {2025},
  month        = {August},
  day          = {11},
  note         = {First published on Notion},
  url          = {https://hijkzzz.notion.site/prorl-v2}
}

@article{hu2025brorl,
  title={Brorl: Scaling reinforcement learning via broadened exploration},
  author={Hu, Jian and Liu, Mingjie and Lu, Ximing and Wu, Fang and Harchaoui, Zaid and Diao, Shizhe and Choi, Yejin and Molchanov, Pavlo and Yang, Jun and Kautz, Jan and others},
  journal={arXiv preprint arXiv:2510.01180},
  year={2025}
}

@article{liu2025part,
  title={Part i: Tricks or traps? a deep dive into rl for llm reasoning},
  author={Liu, Zihe and Liu, Jiashun and He, Yancheng and Wang, Weixun and Liu, Jiaheng and Pan, Ling and Hu, Xinyu and Xiong, Shaopan and Huang, Ju and Hu, Jian and others},
  journal={arXiv preprint arXiv:2508.08221},
  year={2025}
}

@article{song2025fastcurl,
  title={FastCuRL: Curriculum Reinforcement Learning with Stage-wise Context Scaling for Efficient Training R1-like Reasoning Models},
  author={Song, Mingyang and Zheng, Mao and Li, Zheng and Yang, Wenjie and Luo, Xuan and Pan, Yue and Zhang, Feng},
  journal={arXiv preprint arXiv:2503.17287},
  year={2025}
}

@article{setlur2025e3,
  title={e3: Learning to Explore Enables Extrapolation of Test-Time Compute for LLMs},
  author={Setlur, Amrith and Yang, Matthew YR and Snell, Charlie and Greer, Jeremy and Wu, Ian and Smith, Virginia and Simchowitz, Max and Kumar, Aviral},
  journal={arXiv preprint arXiv:2506.09026},
  year={2025}
}

@misc{Polaris2025,
    title = {POLARIS: A Post-Training Recipe for Scaling Reinforcement Learning on Advanced Reasoning Models},
    url = {https://hkunlp.github.io/blog/2025/Polaris},
    author = {An, Chenxin and Xie, Zhihui and Li, Xiaonan and Li, Lei and Zhang, Jun and Gong, Shansan and Zhong, Ming and Xu, Jingjing and Qiu, Xipeng and Wang, Mingxuan and Kong, Lingpeng},
    year = {2025}
}

@article{yu2025dapo,
  title={Dapo: An open-source llm reinforcement learning system at scale},
  author={Yu, Qiying and Zhang, Zheng and Zhu, Ruofei and Yuan, Yufeng and Zuo, Xiaochen and Yue, Yu and Dai, Weinan and Fan, Tiantian and Liu, Gaohong and Liu, Lingjun and others},
  journal={arXiv preprint arXiv:2503.14476},
  year={2025}
}

@article{CompassVerifier,
  title={CompassVerifier: A Unified and Robust Verifier for Large Language Models}, 
  author={Shudong Liu and Hongwei Liu and Junnan Liu and Linchen Xiao and Songyang Gao and Chengqi Lyu and Yuzhe Gu and Wenwei Zhang and Derek F. Wong and Songyang Zhang and Kai Chen},
  year={2025}
}

@inproceedings{sheng2025hybridflow,
  title={Hybridflow: A flexible and efficient rlhf framework},
  author={Sheng, Guangming and Zhang, Chi and Ye, Zilingfeng and Wu, Xibin and Zhang, Wang and Zhang, Ru and Peng, Yanghua and Lin, Haibin and Wu, Chuan},
  booktitle={Proceedings of the Twentieth European Conference on Computer Systems},
  pages={1279--1297},
  year={2025}
}

@article{liu2025dler,
  title={DLER: Doing Length pEnalty Right-Incentivizing More Intelligence per Token via Reinforcement Learning},
  author={Liu, Shih-Yang and Dong, Xin and Lu, Ximing and Diao, Shizhe and Liu, Mingjie and Chen, Min-Hung and Yin, Hongxu and Wang, Yu-Chiang Frank and Cheng, Kwang-Ting and Choi, Yejin and others},
  journal={arXiv preprint arXiv:2510.15110},
  year={2025}
}

@article{li2024numinamath,
  title={Numinamath: The largest public dataset in ai4maths with 860k pairs of competition math problems and solutions},
  author={Li, Jia and Beeching, Edward and Tunstall, Lewis and Lipkin, Ben and Soletskyi, Roman and Huang, Shengyi and Rasul, Kashif and Yu, Longhui and Jiang, Albert Q and Shen, Ziju and others},
  journal={Hugging Face repository},
  volume={13},
  pages={9},
  year={2024}
}

@article{balunovic2025matharena,
  title={Matharena: Evaluating llms on uncontaminated math competitions},
  author={Balunovi{\'c}, Mislav and Dekoninck, Jasper and Petrov, Ivo and Jovanovi{\'c}, Nikola and Vechev, Martin},
  journal={arXiv preprint arXiv:2505.23281},
  year={2025}
}

@article{hendrycks2021measuring,
  title={Measuring mathematical problem solving with the math dataset},
  author={Hendrycks, Dan and Burns, Collin and Kadavath, Saurav and Arora, Akul and Basart, Steven and Tang, Eric and Song, Dawn and Steinhardt, Jacob},
  journal={arXiv preprint arXiv:2103.03874},
  year={2021}
}

@article{lewkowycz2022solving,
  title={Solving quantitative reasoning problems with language models},
  author={Lewkowycz, Aitor and Andreassen, Anders and Dohan, David and Dyer, Ethan and Michalewski, Henryk and Ramasesh, Vinay and Slone, Ambrose and Anil, Cem and Schlag, Imanol and Gutman-Solo, Theo and others},
  journal={Advances in Neural Information Processing Systems},
  volume={35},
  pages={3843--3857},
  year={2022}
}

@inproceedings{he-etal-2024-olympiadbench,
    title = "{O}lympiad{B}ench: A Challenging Benchmark for Promoting {AGI} with Olympiad-Level Bilingual Multimodal Scientific Problems",
    author = "He, Chaoqun  and
      Luo, Renjie  and
      Bai, Yuzhuo  and
      Hu, Shengding  and
      Thai, Zhen  and
      Shen, Junhao  and
      Hu, Jinyi  and
      Han, Xu  and
      Huang, Yujie  and
      Zhang, Yuxiang  and
      Liu, Jie  and
      Qi, Lei  and
      Liu, Zhiyuan  and
      Sun, Maosong",
    editor = "Ku, Lun-Wei  and
      Martins, Andre  and
      Srikumar, Vivek",
    booktitle = "Proceedings of the 62nd Annual Meeting of the Association for Computational Linguistics (Volume 1: Long Papers)",
    month = aug,
    year = "2024",
    address = "Bangkok, Thailand",
    publisher = "Association for Computational Linguistics",
    url = "https://aclanthology.org/2024.acl-long.211/",
    doi = "10.18653/v1/2024.acl-long.211",
    pages = "3828--3850",
}

\appendix
\clearpage




\end{document}